\documentclass{article}
\usepackage{misc/spconf,amsmath,graphicx}

\title{RETHINKING IMAGE HISTOGRAM MATCHING FOR IMAGE CLASSIFICATION}
\name{Rikuto Otsuka, Yuho Shoji, Yuka Ogino, Takahiro Toizumi, and Atsushi Ito}
\address{NEC Corporation}
\usepackage{bm}
\usepackage{url}
\usepackage{color}
\usepackage{amsmath}
\usepackage{amssymb}
\usepackage{amsfonts}
\usepackage{multirow}
\usepackage{algorithm}
\usepackage{algpseudocode}
\usepackage{fancyhdr}
\usepackage{multibib}
\newcites{appx}{References}
\DeclareMathOperator*{\argmin}{arg\,min}
\fancypagestyle{firstpagefooter}{
  \fancyhf{}  
  \fancyfoot[L]{\scriptsize
    © 2025 IEEE. Personal use of this material is permitted. 
    Permission from IEEE must be obtained for all other uses, in any current or future media, 
    including reprinting/republishing this material for advertising or promotional purposes, 
    creating new collective works, for resale or redistribution to servers or lists, 
    or reuse of any copyrighted component of this work in other works.
  }

}

\begin{document}
\maketitle
\thispagestyle{firstpagefooter}

\begin{abstract}
This paper rethinks image histogram matching (HM) and proposes a differentiable and parametric HM preprocessing for a downstream classifier. Convolutional neural networks have demonstrated remarkable achievements in classification tasks. However, they often exhibit degraded performance on low-contrast images captured under adverse weather conditions. To maintain classifier performance under low-contrast images, histogram equalization (HE) is commonly used. HE is a special case of HM using a uniform distribution as a target pixel value distribution. In this paper, we focus on the shape of the target pixel value distribution. Compared to a uniform distribution, a single, well-designed distribution could have potential to improve the performance of the downstream classifier across various adverse weather conditions. Based on this hypothesis, we propose a differentiable and parametric HM that optimizes the target distribution using the loss function of the downstream classifier. This method addresses pixel value imbalances by transforming input images with arbitrary distributions into a target distribution optimized for the classifier. Our HM is trained on only normal weather images using the classifier. Experimental results show that a classifier trained with our proposed HM outperforms conventional preprocessing methods under adverse weather conditions.
\end{abstract}


\section{Introduction}
\label{sec:intro}

\begin{figure*}[t]
  \centering
  \includegraphics[width=\linewidth]{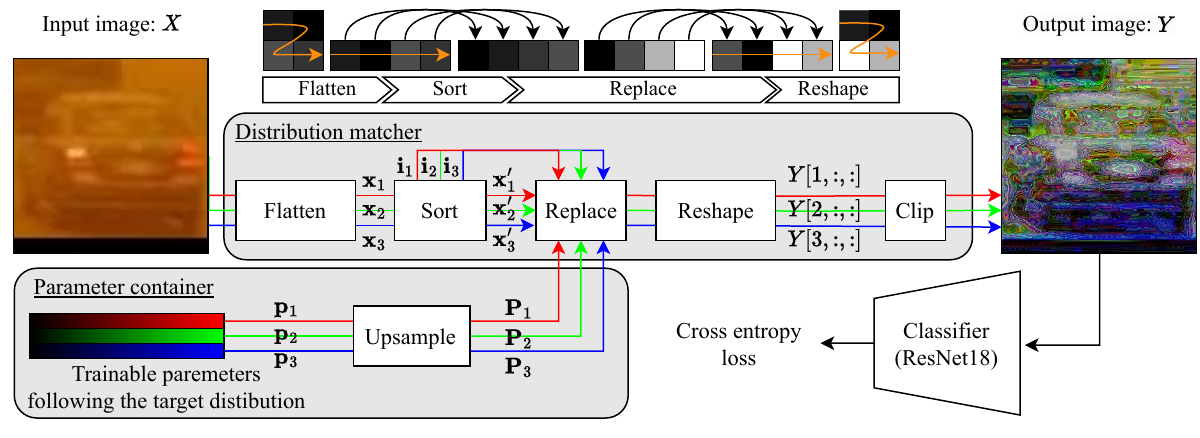}
\caption{Block diagram of the proposed method. The proposed method consists of two main components: a distribution matcher and a parameter container. The distribution matcher transforms the pixel value distribution of input image to match the target distribution. The parameter container stores the trainable parameters $\mathbf{p}_{1}, \mathbf{p}_{2}$, $\mathbf{p}_{3}$ following the target distribution.}
\label{fig:proposed}
\end{figure*}

\begin{figure}[t]
  \centering
  \includegraphics[width=\linewidth]{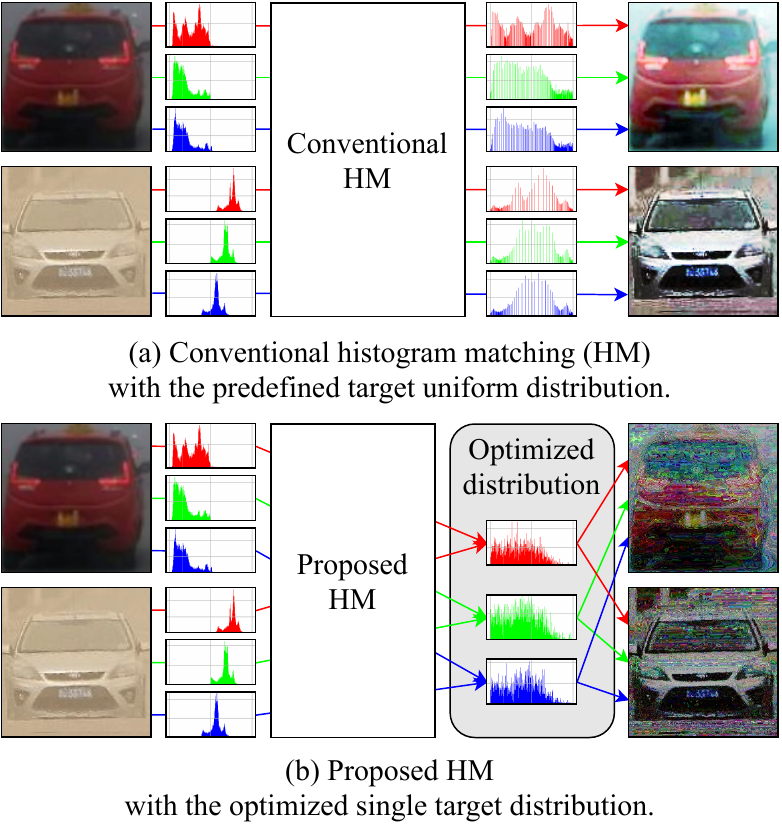}
\caption{Comparison of conventional and proposed histogram matching (HM). (a) Conventional HM with the predefined uniform target distribution. (b) The proposed HM with the optimized single target distribution.}
\label{fig:histogram_is_only_one}
\end{figure}

Convolutional neural networks (CNNs) have achieved remarkable performance in many computer vision applications \cite{2016He, 2021Lengyel}. Most CNNs are trained on images under ideal conditions, such as daytime and clear weather. However, the performance of CNNs often degrades when processing low-contrast images captured under adverse weather conditions, such as low-light, fog, rain, sand, and snow. This limitation poses serious issues for real world applications like autonomous vehicles and surveillance systems. To address these issues, image enhancement methods \cite{2020Guo, 2025Li} have been employed to enhance the contrast of adverse images.

To enhance the contrast of adverse images, tone mapping has been used \cite{2022Rafael}. Tone mapping maps input pixels to output pixels using tone mapping functions. These methods can be categorized into three: look up table-based (LUT-based), curve-based, and histogram-based methods. LUT-based methods use the predefined LUT that maps input pixels to output pixels. Curve-based methods model the tone mapping function using curves, such as gamma curves. Histogram-based methods transform the histogram of an input image by a tone mapping function. In particular, histogram matching (HM) creates a tone mapping function based on the pixel value distribution. This function transforms the pixel value distribution of an input image to the predefined target distribution. A uniform distribution is often used as the target distribution, and this specific case is well known as histogram equalization (HE). However, tone mapping methods often require parameter tuning for each input image. To address this problem, image-adaptive methods have been proposed.

Image-adaptive methods \cite{2020Guo,2022Li,2022Zeng} estimate tone mapping parameters, such as the gamma value \cite{2023Xu} and parameters of LUTs \cite{2022Zeng}, for each input image. These methods improve both visual quality and image recognition performance under adverse weather conditions. To further enhance image recognition performance, image-adaptive models are trained using the loss function of a downstream image recognition model \cite{2023Xu,2022Wang,2024Wang}. However, these models depend heavily on the training dataset and suffer from performance degradation in unknown adverse weather conditions. Therefore, a generalization method is necessary to maintain performance even for unknown adverse weather conditions.

We focus on the potential of HM for generalizing a classifier to unknown adverse weather conditions. HM reduces imbalance of pixel value distribution that degrades recognition performance \cite{2020Wang} by transforming diverse pixel value distributions of input images to a consistent target distribution. We hypothesize that a single well-designed distribution enhances recognition performance across various adverse weather conditions. For this purpose, we aim to optimize this target distribution using the loss function of a downstream classifier. To optimize a target distribution with the downstream classifier, differentiable and parametric HM is necessary. However, unlike the LUT-based and curve-based methods, it is difficult to implement HM in a differentiable and parametric manner \footnote[2]{A discussion on the difficulty is provided in the supplementary materials \url{https://dx.doi.org/10.60864/73sh-mq88}.}. Therefore, we design the HM processing to be differentiable and parametric to train a target distribution.

This paper proposes a differentiable and parametric HM-based image enhancement method. The proposed HM replaces the pixel value distribution of the input image with a target pixel value distribution. Our target distribution is optimized by the loss function of a downstream classifier using images under ideal conditions. Consequently, the optimized target distribution achieves an optimal shape for the downstream classifier. When testing on an adverse weather image, diverse pixel value distributions of any input images are converted to this optimized target distribution. This transformation resolves pixel value imbalance in unknown adverse weather conditions and boosts the performance of the downstream classifier. The experimental results show that the proposed HM generalizes to various adverse weather conditions using only normal weather training data and improves the performance of the downstream classifier.

\section{Related Work}
\label{sec:related}

Tone mapping methods based on LUT, curve, and histogram are generally used to enhance the contrast of images \cite{2022Rafael}. LUT and curve-based methods are particularly studied for image-adaptive methods. These image-adaptive methods use neural networks to estimate tone mapping parameters for each input image, which are then used to enhance image contrast \cite{2020Guo,2022Li,2022Zeng}. ZeroDCE \cite{2020Guo} estimates an image-adaptive and pixel-wise quadratic curve to enhance image contrast. Another study trains multiple 3D LUTs to enhance input images and estimates image-adaptive weights to combine these LUTs \cite{2022Zeng}. While these methods can improve visual quality through image enhancement loss functions, they do not guarantee better performance for image recognition models. To enhance the performance of a image recognition model, some studies have proposed image-adaptive preprocessing for a downstream image recognition model \cite{2023Xu, 2022Wang, 2024Wang, 2024Ogino, 2025Ono, 2023Ljungbergh}. These preprocessing methods have trainable parameters optimized using the loss function of a downstream image recognition model. For instance, in adaptive gamma correction \cite{2023Xu}, a gamma parameter is estimated for each input image to improve image recognition performance. However, to optimize their trainable parameters, these models require training data under adverse weather conditions. This leads to errors when recognizing adverse images not included in the training data.

To recognize adverse images not included in the training data, several zero-shot day-night domain adaptation methods have been proposed \cite{2021Lengyel, 2025Li, 2022Wang, 2024Wang, 2023Luo}. Zero-shot day-night domain adaptation methods can be categorized into two approaches: preprocessing-based methods \cite{2021Lengyel,2022Wang,2024Wang} and training-based methods without preprocessing \cite{2025Li,2023Luo}. Preprocessing-based methods are independent of specific tasks and model architectures. Additionally, preprocessing-based methods can be combined with training-based methods. Given these advantages, we focus on the preprocessing-based methods, and the training-based methods are out of scope. As an example of the preprocessing-based methods, CIConv extracts color-invariant features by replacing the input layer of a CNN to be used like preprocessing \cite{2021Lengyel}.

Histogram matching (HM) transforms the pixel value distribution of an input image to a target distribution \cite{2022Rafael}. Histogram equalization (HE) uses a uniform distribution as the target distribution \cite{2022Rafael}. Contrast limited adaptive HE (CLAHE) \cite{1994Karel} is used not only for contrast enhancement but also as preprocessing for image recognition tasks \cite{2021Lengyel,2020Muhammad}. Additionally, image restoration methods focusing on histogram transformation have been proposed \cite{2019Xiao,2025Sun,2025Peng}. Despite the large potential of HM for generalizing image enhancement to unknown adverse weather conditions, HM has not been used as a trainable preprocessing. This is because of the difficulty in designing HM that is differentiable and trainable. This paper addresses this issue by redesigning HM.

\section{Proposed method}

Fig. \ref{fig:proposed} shows a block diagram of the proposed HM. We prepare trainable parameters following the target distribution and use them to replace all pixel values in the input image. We sort the pixels using their pixel values as keys and maintain their relative ordering during this replacement. Fig. \ref{fig:histogram_is_only_one} illustrates the difference between the conventional HM and the proposed HM. Fig. \ref{fig:histogram_is_only_one} (a) shows the conventional HM with the predefined target uniform distribution, where the output pixel value distributions depend on the input images and differ from each other. In contrast, Fig. \ref{fig:histogram_is_only_one} (b) shows that the proposed HM generates a consistent output distribution across different input images. This distribution is optimized using the loss function of a downstream classifier.

As shown in Fig. \ref{fig:proposed}, our HM consists of two components: a distribution matcher and a parameter container. The distribution matcher matches the pixel value distribution of the input image to the target distribution for each color channel. The parameter container contains the trainable parameters following the target distribution. We represent the pixel value distribution of an input image as a sorted vector of pixels. This representation retains the pixel value distribution while discarding the spatial information of the input image.

To obtain a representation of the input image, we first flatten the 2D image into a 1D vector ${\mathbf x}_c \in \mathbb{R}^{HW}$, for each color channel $c$ of the input image $X \in \mathbb{R}^{C\times H \times W}$. Here, $C$, $H$, and $W$ represent the size of channels, the height, and the width of the image, respectively. We then sort ${\mathbf x}_c$ by its values to obtain the sorted vector ${\mathbf x}'_c$ and its corresponding index ${\mathbf i}_c$. For the representation of the target distribution, we upsample the trainable parameters ${\mathbf p}_c \in {\mathbb R}^{s}$ to ${\mathbf P}_c \in {\mathbb R}^{HW}$ using linear interpolation, where $s$ is the size of trainable parameters. The values in ${\mathbf x}'_c$ are replaced with values from ${\mathbf P}_c$ using the sorted index ${\bf i}_c$, and the result is reshaped to the original image size, yielding ${\mathbf Y} \in {\mathbb R}^{C \times H \times W}$. Finally, the values of the output pixels are clipped to the range from 0 to 1 to constrain pixel values. Through this process, we modify conventional HM to be differentiable. Moreover, our HM allows for setting the target distribution to arbitrary distribution.

We then input the transformed image ${\bf Y}$ into a classifier $F_\theta$ with parameters $\theta$. The loss function is the cross entropy loss between the classifier output and the ground truth label. During the training, we optimize both $\theta$ and ${\bf p}_1, \ldots, {\bf p}_C$.

\section{Experiments}
\label{sec:majhead}

\begin{table*}[t]
    \centering
    \caption{Results of top-1 accuracy comparison. The \textbf{best} value is highlighted by bold. The adverse mean column shows the average values for adverse weather conditions (night, fog, rain, sand, and snow). The fixed in the proposed category is the result when trainable parameters $\mathbf{p}_1, \mathbf{p}_2, \mathbf{p}_3$ are fixed ($\mathbf{p}_1, \mathbf{p}_2, \mathbf{p}_3$ are following a uniform distribution).}
    \begin{tabular}{ll | c | ccccc | c} 
        \hline
        Category & Method & day & night & fog & rain & sand & snow & adverse mean\\
        \hline
        Baseline & - & 79.80 & 49.98 & 14.29 & 19.80 & 17.51 & 29.32 & 35.12\\
        \hline
        LUT-based & Zeng et al. \cite{2022Zeng} & 78.80 & 53.04 & 17.05 & 22.59 & 18.55 & 29.52 & 28.15\\
        \hline
        \multirow{3}{*}{Curve-based}
        & ZeroDCE \cite{2020Guo} & 76.28 & 58.75 & 7.87 & 13.50 & 12.67 & 14.38 & 21.43\\
        & Ljungbergh et al. \cite{2023Ljungbergh} & 80.64 & 51.51 & 14.98 & 20.05 & 19.45 & 23.97 & 25.99\\
        & Xu et al. \cite{2023Xu} & 80.12 & 55.65 & 17.12 & 21.03 & 23.55 & 22.12 & 27.89\\
        \hline
        \multirow{1}{*}{Domain adaptation}
        & CIConv \cite{2021Lengyel} & 80.80 & \bf{60.95} & 19.46 & 25.61 & 31.52 & 23.97 & 32.30\\
        \hline
        \multirow{3}{*}{Histogram-based}
        & HE \cite{2022Rafael} & 79.92 & 59.02 & 31.54 & 32.57 & 49.78 & 31.30 & 40.84\\
        & CLAHE \cite{1994Karel} & \bf{82.76} & 58.52 & 38.51 & 42.06 & 56.63 & 38.49 & 46.84\\
        & Histoformer \cite{2025Sun} & 79.72 & 49.62 & 15.32 & 20.13 & 17.66 & 34.25 & 27.40\\
        \hline
        \multirow{2}{*}{Proposed}
        & Fixed (uniform distribution) & 79.96 & 59.69 & 34.30 & 36.50 & 52.83 & 31.58 & 42.98\\
        & Trainable & 77.56 & 57.35 & \bf{44.24} & \bf{47.22} & \bf{68.55} & \bf{45.27} & \bf{52.53}\\
        \hline
    \end{tabular}
    \label{tab:performance_comparison}
\end{table*}

\begin{figure}[t]
  \centering
  \includegraphics[width=\linewidth]{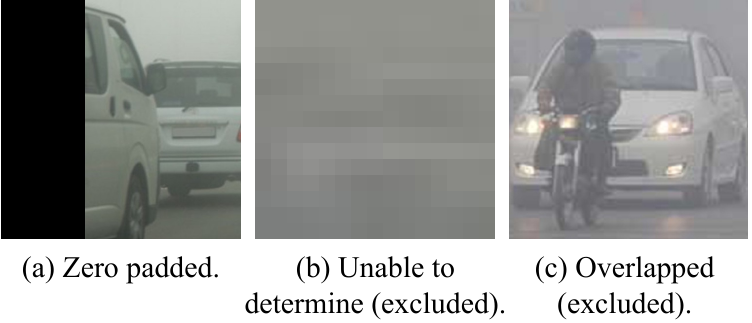}
  \caption{Examples of zero padded and excluded images. (a) An edge cropped image with zero padding. (b) A cropped image unable to determine the object class manually. (c) A cropped image with multiple overlapped objects of different classes.}
\label{fig:dawn}
\end{figure}

To evaluate the performance of the proposed method in adverse weather conditions, we used two datasets: CODaN \cite{2021Lengyel} and a modified version of DAWN \cite{2020Mourad}. The CODaN dataset is designed for classification and contains 10 classes of daytime and nighttime images. This dataset consists of cropped images from object detection datasets. CODaN includes 10,000 daytime training images, 500 daytime validation images, 3,000 daytime test images, and 3,000 nighttime test images. The DAWN dataset was originally created for object detection under adverse weather conditions including fog, rain, sand, and snow. We modified the DAWN dataset for classification, following the way of the CODaN dataset. We used all classes from CODaN, while from DAWN we selected four classes: 'bicycle', 'car', 'motorcycle', and 'bus'. These classes are common to both datasets. We cropped instances of these object classes from the original images using bounding box information. We created a square crop with sides equal to the longer edge of the bounding box. The center position of the cropped square image is aligned on that of its original bounding box. We finally resized these crops to $224\times224$ pixels. Any areas without pixels were zero padded. We excluded images where the object class was unable to determine, and those with multiple overlapped objects. Fig. \ref{fig:dawn} shows examples of zero-padded and excluded images. Our modified DAWN dataset consists of 1,449 fog images, 1,222 rain images, 1,342 sand images, and 1,460 snow images.

For training, we utilized the daytime training set of the CODaN dataset. The proposed method and classifier were trained end-to-end. The size $s$ of trainable parameters ${\bf p}_c$ was set to $2048$, and their initial values were set to linearly increasing values from 0 to 1. We employed ResNet18 \cite{2016He} as the backbone of the classifier $F_\theta$ and configured the training settings based on prior research \cite{2021Lengyel}. The weight parameters of ResNet18 were initialized with random values. We used stochastic gradient descent (SGD) as the optimizer with an initial learning rate of 0.05, weight decay of 0.0001, and momentum of 0.9. The learning rate was reduced to 0.005 and 0.0005 at 50 and 100 epochs, respectively. We set the batch size to 64 and trained for a total of 175 epochs. For data augmentation, we employed random horizontal flips, random cropping, random rotations, and color jitter. For evaluation, we used test images from the CODaN dataset (daytime, nighttime) and the DAWN dataset (fog, rain, sand, snow).

\begin{figure*}[t]
  \centering
  \includegraphics[width=\linewidth]{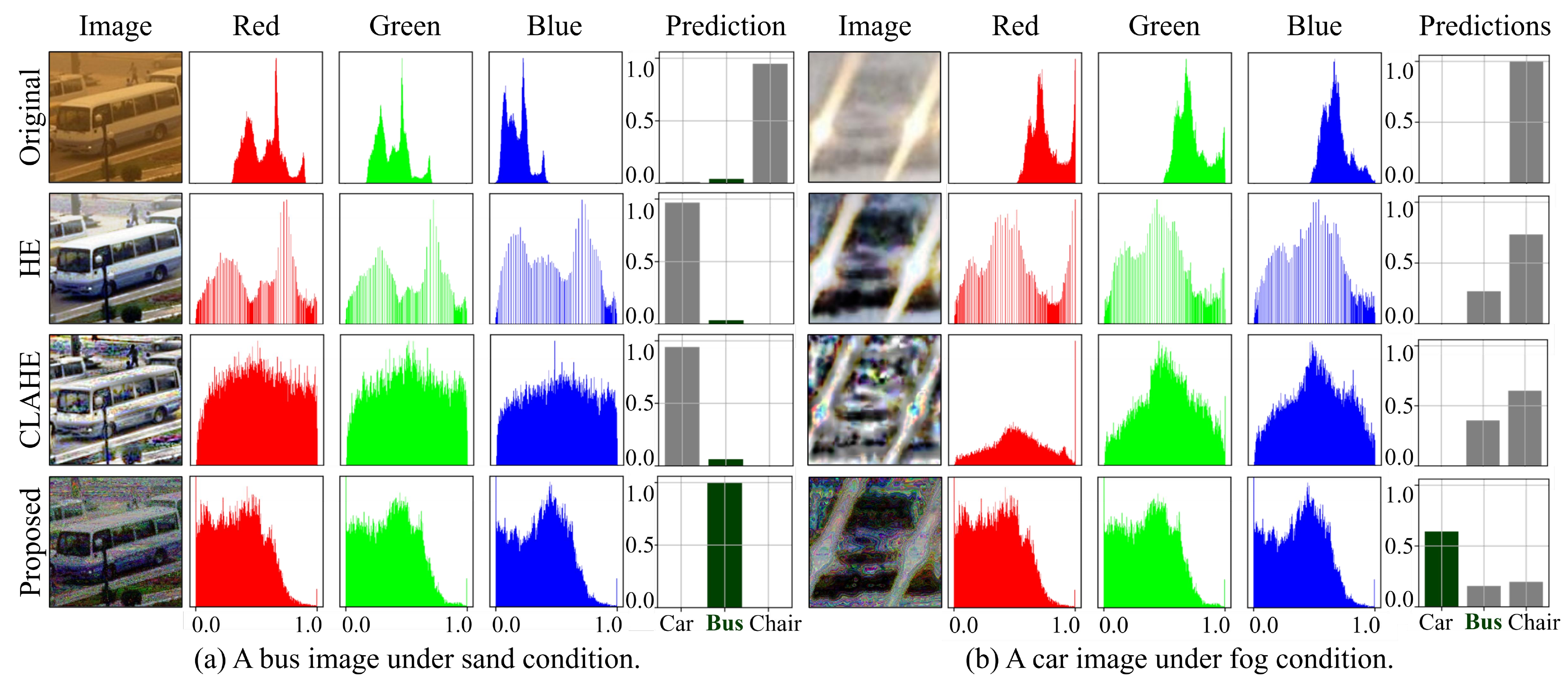}
\caption{Examples of images and histograms transformed by conventional histogram-based method and proposed method.}
\label{fig:hist_result}
\end{figure*}

We compared top-1 accuracy of the proposed method with conventional methods. Since the proposed method is categorized as a preprocessing method, we selected conventional preprocessing methods for comparison. Other methods are out of scope because these methods can be used with preprocessing methods. The selected conventional methods include LUT-based \cite{2022Zeng}, curve-based tone mapping \cite{2020Guo, 2023Xu, 2023Ljungbergh}, domain adaptation \cite{2021Lengyel}, and histogram-based methods \cite{2022Rafael, 1994Karel, 2025Sun} To investigate performance degradation under adverse weather conditions, we used a baseline trained on the daytime training images of the CODaN dataset without preprocessing. LUT-based, Curve-based tone mapping and domain adaptation were employed to examine how adaptation to one known condition affects classifier performance on unseen adverse weather conditions. Histogram-based methods were employed to confirm the potential for generalizing classifiers to unseen adverse weather conditions.

Table \ref{tab:performance_comparison} compares conventional methods and the proposed method. The baseline showed significant performance degradation under night, fog, rain, sand, and snow. Curve-based tone mapping and domain adaptation demonstrated improvements in specific target domains like night, in addition to day. However, the improvement in fog, rain, sand, and snow tended to be smaller compared to the night. HE and CLAHE showed performance improvements across various adverse weather conditions compared to other conventional methods. We confirmed that histogram-based methods improve their classification performance even under unseen adverse weather conditions. This result also indicates the generalization ability of the histogram-based methods. When the initial values of the trainable parameters were fixed to a uniform distribution, the proposed method achieved performance equivalent to HE. Among histogram-based methods, the proposed method with trainable parameters achieves further improvements under fog, rain, sand, and snow conditions, at the expense of a slight reduction in day performance. This is because the proposed method optimizes the target distribution for classifier. These results confirm that the proposed method maintains the generalization ability of histogram-based methods while improving performance on the classification task.

\begin{table}[tbp]
    \centering
        \caption{Comparison of computation times for each method.}
        \begin{tabular}{l | r | r}
            \hline
            Method & CPU time [ms] & GPU time [ms]\\
            \hline
            Histoformer \cite{2025Sun} & 4519.5 & 97.03 \\
            CIConv \cite{2021Lengyel} & 131.2 & 0.45 \\
            ZeroDCE \cite{2020Guo} & 59.7 & 0.75 \\
            ResNet18 \cite{2016He} & 30.1 & 0.82 \\
            Xu et al. \cite{2023Xu} & 22.4 & 0.73 \\
            Ljungbergh et al. \cite{2023Ljungbergh} & 0.5 & 0.02\\
            \hline
            Proposed & 1.5 & 0.19 \\
            \hline
        \end{tabular}
    \label{tab:cpu_gpu_times}
\end{table}

Fig. \ref{fig:hist_result} shows the differences among HE, CLAHE, and the proposed method. The left and right blocks show examples of transformed images under different scenes. For each block, from left to right, images show the transformed image, histograms for each color channel, and the classification results. For the transformed images, HE enhanced the global contrast, while CLAHE enhanced local contrast. However, the proposed method converted the input image to a noisy image. This noise occurs because no loss function is used to enhance visual quality, and this noise is a limitation of our method. For the transformed histograms, conventional HE preserved the features of the original histogram. CLAHE converted distributions to more uniform distributions than those of HE. However, although the initial distribution was uniform, the proposed method significantly transformed the distribution into a non-uniform distribution, suggesting that uniformity might not be optimal for the classifier.

In the classification results, dark green denotes the ground truth (GT) class. While the original, HE, and CLAHE images failed to classify the image correctly, the proposed method successfully identified the GT class. These results suggest that the proposed method enhances classification results by optimizing the pixel value distribution, while reducing visual quality.

We additionally evaluated the computation time of the conventional methods and the proposed method. Computation times were measured on a CPU (Intel(R) Core(TM) i9-14900) using a single thread and on a GPU (NVIDIA GeForce RTX3080). The times are measured as averages of 10 executions. For fair comparison, we only compared methods implemented in PyTorch. Table \ref{tab:cpu_gpu_times} presents the computation times for each method. We included the processing time of ResNet18 as the downstream classifier for comparison. As shown in Table \ref{tab:cpu_gpu_times}, we confirmed that the proposed method is significantly faster than other conventional methods, except for Ljungbergh et al. \cite{2023Ljungbergh}. Moreover, it achieves a computation time substantially lower than that of the ResNet18. These results demonstrate that our proposed method does not introduce significant overhead compared to ResNet18.

\section{Conclusion}
\label{sec:conclusion}
This paper proposed an optimization method for image pixel value distributions to enhance the performance of downstream classifiers. We redesigned the HM to be differentiable and parametric. The proposed method can optimize the target distribution using a loss function of the downstream classifier. Experimental results confirmed that the proposed method can train a target distribution different from conventional HM. Furthermore, the results also showed that the proposed method generalizes to various adverse weather conditions using only normal weather training data and improves the performance of the downstream classifier.

\newpage
\appendix
\setcounter{figure}{4}
\setcounter{table}{2}
\setcounter{algorithm}{0}

\maketitlesupplementary 

\section{Differentiability of Conventional Histogram Matching}
We explain why it is difficult to modify conventional histogram matching (HM) to be differentiable. This challenge is due to conventional HM containing two non-differentiable processes. We first explain the processes of conventional HM. We then explain the two non-differentiable processes in conventional HM: (1) cumulative distribution function (CDF) calculation and (2) look up table (LUT) calculation.

\subsection{Processes of Conventional Histogram Matching}
Conventional HM \cite{2022Rafael} transforms the CDF of a source image to match the CDF of a target image. Conventional HM first computes the CDFs of both the source image $\mathbf{x}_S$ and the target image $\mathbf{x}_T$, for each color channel. We denote the CDF of $\mathbf{x}_{S}$ as $F_{\mathbf{x}_{S}}$ and the CDF of $\mathbf{x}_{T}$ as $F_{\mathbf{x}_{T}}$. Then, conventional HM creates a mapping function $M(p)$ of pixel value $p$ to match $F_{\mathbf{x}_{S}}$ to $F_{\mathbf{x}_{T}}$. Theoretically, the mapping function $M(p)$ is expressed as $F_{\mathbf{x}_{T}}^{-1}(F_{\mathbf{x}_{S}}(p))$. However, in general, finding analytical expressions for $F_{\mathbf{x}_{T}}^{-1}$ is not a trivial task \cite{2022Rafael}. To address this, in practice, conventional HM uses an approximation method. The mapping function $M(p)$ is expressed as an LUT. Finally, the pixels of the source image $\mathbf{x}_{S}$ with value $p$ are replaced with $M(p)$ to obtain $\mathbf{x}_M$ whose CDF is approximately equal to $F_{\mathbf{x}_{T}}$. Conventional HM involves two non-differentiable processes: (1) CDF calculation and (2) LUT calculation. We explain these processes in the following.

\subsection{Non-Differentiable CDF Calculation}

$F_{\mathbf{x}_T}$ and $F_{\mathbf{x}_S}$ are obtained as the cumulative sums of the histograms of $\mathbf{x}_T$ and $\mathbf{x}_S$. Let $\mathbf{x} = \{x_{i} \mid i = 1, 2, \cdots, H\times W\}$ denote the input $N$-bit digital image flattened to 1D, where $H$ and $W$ represent the height and width of the input image. Let $p \in \mathcal{P} = \{0, 1, \cdots, 2^{N}-1\}$ denote the pixel value of the input image. The histogram of the input image $h_{\mathbf{x}}(p)$ is calculated using the bin assignment function $\delta(x_{i}, p)$:
\begin{align}
h_{\mathbf{x}}(p) & = \sum_{i=1}^{H\times W} \delta(x_i, p),
\end{align}
where, the bin assignment function $\delta(x_{i}, p)$ is defined as:
\begin{align}
\delta(x_{i}, p) = 
\begin{cases}
    1 & \text{if $x_{i} = p$}, \\
    0 & \text{otherwise}.
\end{cases}
\end{align}
For each pixel value $p \in \mathcal{P}$, the CDF $F_{\mathbf{x}}$ is defined as:
\begin{align}
F_{\mathbf{x}}(p) = \frac{1}{HW} \sum_{k=0}^{p} h_{\mathbf{x}}(k),
\end{align}
The derivative of the bin assignment function $\delta(x_i, p)$ w.r.t. $x_{i}$ is zero for all $i$. 
This feature makes the CDF calculation in HM non-differentiable. PyTorch \cite{2017Adam}, a widely used framework in deep learning research, provides histogram calculation functions such as \texttt{torch.histc()}, \texttt{torch.histogram()}, and \texttt{torch.bincount()}. However, these functions lack implementation of a backward function for floating-point inputs\footnote{This behavior was confirmed using PyTorch 2.5.0+cu124.}. To address this zero gradient problem, some researchers \cite{2016Evgeniya,2016Wang,2020Ibrahim,2023Mor} make the histogram calculation differentiable by replacing the bin assignment function with another differentiable function.

\begin{figure}[t]
  \centering
  \includegraphics[width=\linewidth]{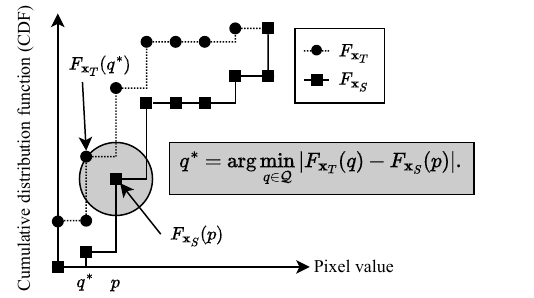}
\caption{The LUT calculation by finding the pixel value $q^{*}$ such that $F_{\mathbf{x}_{T}}(q^{*})$ is the nearest value to $F_{\mathbf{x}_{S}}(p)$.}
\label{fig:hm}
\end{figure}

\begin{algorithm}[t]
\caption{Practical LUT calculation process in HM}
\label{alg:lut}
\begin{algorithmic}[1]
\Require $F_{\mathbf{x}_{S}}$, $F_{\mathbf{x}_{T}}$,  $\mathcal{P} = \mathcal{Q} = \{0, 1,\cdots, 2^{N}-1\}$
\Ensure LUT $M$
\For{each $p \in \mathcal{P}$}
    \State $d_{min} \gets \infty$
    \State $q^* \gets 0$
    \For{each $q \in \mathcal{Q}$}
        \State $d \gets |F_{\mathbf{x}_T}(q) - F_{\mathbf{x}_S}(p)|$
        \If{$d < d_{min}$}
            \State $d_{min} \gets d$
            \State $q^* \gets q$
        \EndIf
    \EndFor
    \State $M(p) \gets q^*$
\EndFor
\State \Return $M$
\end{algorithmic}
\end{algorithm}

\begin{table*}[t]
    \centering
    \caption{Top-1 accuracy (\%) under different size of trainable prameters $s$. $s = 50176$ corresponds to the input size of ResNet18.}
    \label{tab:ablation_param_size}
    \begin{tabular}{c | r | c | ccccc | c}
        \hline
        Category & $s$ & day & night & fog & rain & sand & snow & adverse mean \\
        \hline
        Baseline & - & 79.80 & 49.98 & 14.29 & 19.80 & 17.51 & 29.32 & 35.12\\
        \hline
        \multirow{6}{*}{Proposed} & 256 & 75.32 & 52.90 & 28.99 & 29.30 & 51.71 & 37.60 & 40.10 \\
        & 512 & 75.12 & 54.97 & 32.37 & 33.39 & 53.13 & 40.34 & 42.84 \\
        & 1024 & 75.96 & 56.23 & 39.48 & 39.03 & 61.77 & 39.11 & 47.12 \\
        & 2048 & 77.56 & 57.35 & 44.24 & 47.22 & 68.55 & 45.27 & 52.53 \\
        & 4096 & 77.68 & 58.75 & 39.06 & 37.40 & 60.43 & 37.47 & 46.62 \\
        \hline
        & 50176 & 80.32 & 60.91 & 34.16 & 31.51 & 50.45 & 29.73 & 41.35 \\
        \hline
    \end{tabular}
\end{table*}

\subsection{Non-Differentiable LUT Calculation}
Conventional HM calculates an LUT $M(p)$ using the CDF of the target image $F_{\mathbf{x}_{T}}$ and the CDF of the source image $F_{\mathbf{x}_{S}}$. Then the conventional HM applies the LUT to the source image. Fig. \ref{fig:hm} shows the process of LUT calculation. The LUT of pixel value $M(p)$ is calculated by finding the pixel value $q^{*}$ such that $F_{\mathbf{x}_{T}}(q^{*})$ is the nearest value to $F_{\mathbf{x}_{S}}(p)$. Specifically, $q^*$ is obtained as:
\begin{align}
q^{*} = \argmin_{q \in \mathcal{Q}} |F_{\mathbf{x}_{T}}(q) - F_{\mathbf{x}_{S}}(p)|. \label{eq:argmin}
\end{align}
where $\mathcal{Q} = \{0, 1,\cdots, 2^{N}-1\}$ is the set of possible pixel values (domain of the $F_{\mathbf{x}_{T}}$). For each pixel value $p \in \mathcal{P}$, LUT $M(p)$ is defined as:
\begin{align}
M(p) = q^*
\end{align}
The optimal $q^{*}$ is given by $F_{\mathbf{x}_{T}}^{-1}(F_{\mathbf{x}_{S}}(p))$. However, in general, finding analytical expressions for $F_{\mathbf{x}_{T}}^{-1}$ is not a trivial task \cite{2022Rafael}. In practice, to obtain $q^{*}$ for each $p \in \mathcal{P}$, conventional HM calculate $|F_{\mathbf{x}_{T}}(q) - F_{\mathbf{x}_{S}}(p)|$ for each $q \in \mathcal{Q}$ and find the $q$ that minimizes $|F_{\mathbf{x}_{T}}(q) - F_{\mathbf{x}_{S}}(p)|$. Algorithm \ref{alg:lut} shows this practical LUT calculation process. In this practical LUT calculation, the derivative of $M(p)$ w.r.t. $F_{\mathbf{x}_{T}}(k)$ is zero for all $k \in \mathcal{P}$. This is because $M(p)$ is not a function of $F_{\mathbf{x}_{T}}(k)$, for all $k$. This feature makes the LUT calculation in HM non-differentiable.

\section{Ablation Study on Size of Trainable Parameters}
We present an ablation study on the size of trainable parameters $s$, which was not included in the main paper due to page limitations. We set the minimum value of $s$ to $2^{8}$, corresponding to the bit depth of the images used in our experiments, and the maximum value to $50176$ ($224 \times 224$), corresponding to the input size of ResNet18. The intermediate values were selected as $2^{9}$, $2^{10}$, $2^{11}$, and $2^{12}$. 

Table \ref{tab:ablation_param_size} shows the comparison results of top-1 accuracy under different $s$ values. A result of the baseline trained on the daytime subset of the CODaN dataset without any preprocessing are reprinted for comparison. Under adverse weather conditions, all examined parameter settings achieved higher top-1 accuracy than the baseline.

\bibliographystyle{misc/IEEEbib}
\bibliography{refs/main}
\end{document}